\documentclass[english]{lni}

\usepackage{graphicx}
\usepackage{fancyhdr}
\usepackage{changepage} 
\usepackage{listings} 

\usepackage[figurename=Fig., tablename=Tab., small]{caption}

\fancypagestyle{titlepage}{
\fancyhead[RO]{\small F.  Boutros,  N. Damer,  M. Fang,  M. Gomez-Barrero,  K. Raja,  \linebreak C. Rathgeb,   A.  Sequeira,  and M. Todisco (Eds.): BIOSIG 2024, \linebreak Lecture Notes in Informatics (LNI), Gesellschaft f\"ur Informatik, Bonn 2024} 
\fancyfoot{}}

\renewcommand{\headrulewidth}{0.4pt} 
\author{Afzal Hossain\footnote{ PhD Student, Electrical and Computer Engineering, Clarkson University, 8 Clarkson Avenue, Potsdam, NY13676, US, afhossa@clarkson.edu} \, Tipu Sultan\footnote{PhD Student, Mechanical and Aerospace Engineering, Clarkson University, 8 Clarkson Avenue, Potsdam, NY13676, US, sultant@clarkson.edu} \, Stephanie Schuckers\footnote{Professor, Electrical and Computer Engineering, Clarkson University, 8 Clarkson Avenue, Potsdam, NY13676, US, sschucke@clarkson.edu}}
\title{Deep Learning Approach for Ear Recognition and Longitudinal Evaluation  in Children}

\begin{document}

\maketitle

\renewcommand{\refname}{References}
\setcounter{footnote}{2} 
\thispagestyle{titlepage}
\pagestyle{fancy}
\fancyhead{} 
\fancyhead[RO]{\small DL Approach for Ear Recognition and Longitudinal Evaluation of Ear Recognition in Children \hspace{25pt}  \hspace{0.05cm}}
\fancyhead[LE]{\hspace{0.05cm}\small  \hspace{25pt} Afzal Hossain, Tipu Sultan, and Stephanie Schuckers}
\fancyfoot{} 
\fancyfoot[C]{\thepage} 
\renewcommand{\headrulewidth}{0.4pt} 

\begin{abstract}
Ear recognition as a biometric modality is becoming increasingly popular, with promising broader application areas. While current applications involve adults, one of the challenges in ear recognition for children is the rapid structural changes in the ear as they age. This work introduces a foundational longitudinal dataset collected from children aged 4 to 14 years over a 2.5-year period and evaluates ear recognition performance in this demographic. We present a deep learning based approach for ear recognition, using an ensemble of VGG16 and MobileNet, focusing on both adult and child datasets, with an emphasis on longitudinal evaluation for children.
\end{abstract}

\section{Introduction}
Biometrics involves identifying a person based on their inherent physiological or behavioral characteristics. Physiological biometrics rely on measurements of external physical traits, including fingerprint, iris, face features, and ear structure. Behavioral biometrics typically assess learned behaviors, such as gait, respiration pattern, keyboard typing pattern, handwriting, speech, and heartbeat. Several systems have been developed utilizing these biometric traits and evaluated in real-world scenarios. Among the various physiological biometric traits, the ear has gained significant attention in recent years due to its proven reliability for human recognition and other security applications\cite{AB01}. Ear recognition presents a viable alternative to more conventional biometric methods \cite{AB02} , \cite{AB03}. Ear biometrics offer advantages such as reduced invasiveness during capture and less control requirements during image acquisition compared to other modalities. Human ear possesses several other desirable characteristics such as capability to differentiate between identical twins, and insensitivity to emotions and facial expressions \cite{AB04}, \cite{AB05}, \cite{AB06}. Additionally, it is reasonable to argue that there are fewer privacy implications associated with ear recognition in contrast to facial recognition. With these advantages, we can construct and refine reliable ear recognition systems on various devices in a non-invasive and unobtrusive manner \cite{AB07}, \cite{AB08}.

The most effective biometric modalities utilized in various security applications include iris recognition, fingerprint recognition, and facial recognition. The ear biometric system can supplement other biometric systems and offer identity clues when the information from the other systems is inconsistent or even absent. In the case of capturing images, it is simple to take ear images, similar to face. The human ear is visible even when a mask is worn (especially during an epidemic time like Covid) where the face is not fully visible. In the context of the COVID-19 scenario, fingerprint and palmprint-based recognition are even not suitable due to their reliance on contact-based feature extraction. Iris-based recognition systems are expensive due to the specialized sensors necessary for feature extraction from the iris. Additionally, the biometric systems discussed above necessitate the active participation and cooperation of individuals for accurate identification. However, using these systems in busy places like train stations, museums, and malls is hard because there are so many people and it can get crowded. In such crowded settings, traditional biometric methods that require physical contact or the voluntary cooperation of individuals become impractical and inefficient. Consequently, there arises a pressing need for alternative biometric solutions that are contactless and non-cooperative. Ear biometrics emerges as a promising solution to address this need.

While ear biometrics offer numerous advantages, including reduced invasiveness and enhanced privacy compared to other modalities, it's crucial to understand the stability and developmental stages of the ear. The ear exhibits long-term stability, however, there are periods during which the ear undergoes rapid changes. Medical literature indicates that during the initial four months after birth, the ear undergoes proportional growth in all dimensions, followed by a gradual increase in size thereafter \cite{AB09}. The forensic science literature indicates that significant alterations in ear shape occur between the ages of four months and eight years where the rate of elongation is approximately five times greater in the period. Following this period, its dimensions remain relatively constant until approximately 70 years of age, after which there may be a resurgence in size towards the ear lobe \cite{AB10}. Thus, prior to the age of eight, the ear undergoes significant changes, rendering it unstable, whereas beyond this age, its structure stabilizes. This understanding of ear development informs the approach taken in employing advanced models for ear recognition. Early research in ear recognition showed significant improvement in performance, where the methods typically involved manual feature engineering to describe important ear features, which were then used to train traditional classifiers. However, these techniques faced limitations due to the need for expertise in feature extraction and the time-consuming nature of manual methods. Additionally, performance suffered when faced with variability in image appearance. In recent years, deep learning (DL) algorithms have brought about significant advancements in various application domains, including biometric recognition \cite{AB11}. DL models streamline the process by performing both feature extraction and classification in an end-to-end manner, eliminating the need for manual feature extraction. To segment ears from images, we employ Mask R-CNN, followed by feature extraction MobileNet model. Euclidean distances between these features are then calculated to determine recognition performance. We evaluate our child ear dataset, collected from local schools, focuses on children aged between 4 and 14 years. This dataset includes images of children under 8 years old, during which the ear structure undergoes significant changes, as well as images of children above 8 years old, when the ear structure becomes stable.

\section{Related Works}

\begin{table}[htb]
\centering
\begin{tabular}{l l l l l l l}
\hline
Datasets & Subjects & Samples  & Ages & Time Gap & Sessions \\
\hline
Molla et al. \cite{AB30} & 71 & 446   & 0yrs - 1yrs & - & 1\\
Tiwari et al. \cite{AB32} & 210 & 1027  & 0hrs - 48hrs & - & 1\\
Ntshangase et al. \cite{AB31} & 100 & 1000  & 0days - few mnths & - & 1\\
Etter et al. \cite{AB43} & 224 & 896 & 6days - 6mnths & - & 1\\
Child Ear Data (ours) & 231 & 2770  & 3yrs – 18yrs & 6mnths & 6\\
\hline
\end{tabular}
\caption{Comparison of Ear Recognition Datasets Used in Child Ear Recognition Research}
\label{tab1}
\end{table}

Ear recognition for children was initially pioneered in 1960 by Fields et al. \cite{AB29}. Their seminal work involved the manual analysis of the ears of newborns from a database comprising 206 participants. Recognizing the challenges associated with accurately identifying children, the authors explored potential strategies for utilizing ear features to distinguish between newborns. Fields et al. \cite{AB29} ultimately affirmed that visual examination of ears could serve as a viable method for distinguishing among children. To the best of our knowledge, there are currently no commercially available automated systems specifically designed for ear recognition in children \cite{AB30}, \cite{AB31}. Automated ear recognition in children still remains an active area of research. However, recently some efforts have been made in this direction. In 2015, Tiwari et al. \cite{AB32} introduced a fully automated ear recognition system for newborns. Their approach involved automatically locating, segmenting, and cropping the ear region in the provided ear image. They also explored a novel method for automatically recognizing newborns. The authors demonstrated that their algorithm offers a computationally efficient solution for automatic newborn recognition, achieving an identification accuracy of 89.28\% on a database of 210 subjects. An article published in October 2017 mentioned that the MATLAB Health Research Centre in Bangladesh and the Angkor Hospital for Children in Cambodia will collaborate to evaluate various biometric modalities, including fingerprints, irises, palm prints, ears, and feet, to determine which is most suitable for infants and young children \cite{AB33}. Tiwari first investigated if automated ear recognition of newborns can be done in 2011 \cite{AB34}. Subsequently in \cite{AB32}, Tiwari introduced an enhancement to ear recognition for newborns by integrating ear features with soft biometrics and in \cite{AB35} a multimodal database of newborns was compiled for biometric recognition incorporating soft biometrics. In 2016, Tiwari et al. \cite{AB36} examined adult and infant ear images to explore automated identification utilizing 2D ear images. This study involved the comparison of seven ear recognition algorithms using a dataset comprising both adult and infant subjects. In May 2017, an ongoing project was reported involving the Council for Scientific and Industrial Research (CSIR) in South Africa. They are in the process of developing a biometric system capable of identifying or verifying the identities of children from infancy through childhood. CSIR researchers will evaluate three biometric modalities: fingerprints, iris patterns, and ear shape, to determine which of the three is best suited for the system \cite{AB37}. However, the challenges associated with ear recognition for children are well-documented, including geometric changes due to growth and variations in illumination during image acquisition. While efforts have been made to address these challenges, existing works have primarily focused on databases consisting of adults. Table 1 presents an overview of key studies conducted in the field of ear recognition for infants and children.

For child ear segmentation, we utilize Mask R-CNN, followed by feature extraction using the MobileNet model. Euclidean distances between these features are then computed to assess recognition performance. Our child ear dataset, gathered from local schools where majority of them are aged between 4 and 14 years. The remainder of the paper is structured as follows:

In Section 3, we present an overview of the segmentation model and the pretrained model utilized in our proposed architecture. Section 4 provides comprehensive details on the child ear database along with ear image pre-processing steps, training and evaluation procedure, and experimental results with discussions. Finally, Section 5 summarizes our study.

\section{Network Architecture}

In our ear recognition method, we first mask the ear region from profile face images, creating JSON files with ear coordinates. We utilize Mask R-CNN  \cite{AB38} for ear segmentation, using its multi-task loss function and RoIAlign layer to optimize mask prediction accuracy. Using the Matterport Mask R-CNN library and TensorFlow, we train the model on our dataset of annotated ear images. A framework of Mask R-CNN is shown in Figure 1. For feature extraction, we ensemble features from both VGG16  \cite{AB39} and MobileNet  \cite{AB40} models, combining them into a single feature vector and apply t-SNE. We then compute Euclidean Distances between these feature vectors to generate matching scores, which are used to calculate the True Acceptance Rate (TAR) and False Acceptance Rate (FAR), enhancing the accuracy of our ear recognition system.

\begin{figure}[htb]
\centering
    \includegraphics[width=6cm]{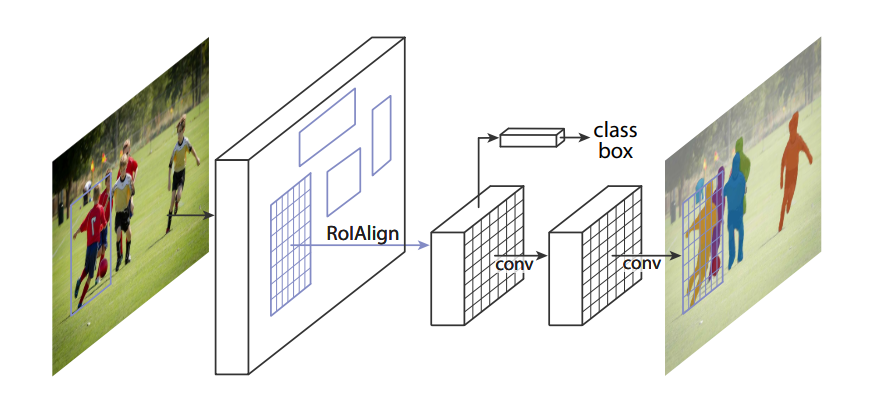}
    \caption{The Mask R-CNN framework for instance segmentation}
    \label{logo}
\end{figure}

\section{Experiment}

\subsection{Dataset and Pre-processing of Data}
For our experimental purposes, our research team partners with a nearby elementary, middle, and high school to identify and recruit participants who voluntarily agree to take part, following an approved Institutional Review Board (IRB) protocol. We capture the profile face images with ear using a DSLR camera at a resolution of 3648 by 5472 pixels. Image acquisition is conducted under controlled indoor lighting conditions and minimal variation in pose. Our dataset comprises 231 subjects (on average 2samples per subject) collected in a controlled environment over a period of 3 years, with a time-lapse of 6 months between sessions. Therefore, we have 6 collections of child dataset. However, participation was not consistent across all sessions, as some subjects did not participate in every session. The fluctuation in participant numbers per session is influenced by factors such as newly enrolled participants each school year, absenteeism, unwillingness to participate on certain days, and participants relocating out of the school district. Out of the total, 209 subjects participated in more than one session. Our dataset exhibits gender balance, with 117 female and 114 male subjects. Enrollment age ranges from 3 years to 18 years, with the majority falling between 4 and 14 years. To the best of our knowledge, there is no other available dataset of child ear images spanning such an extended period. Our dataset is available for research purposes upon request via email.

After applying masking to the images, we isolate only the ear portions while turning the background black. Subsequently, we iterate over the images, align and rotate each image, and remove black border to ensure tight cropping. We flatten each image and resize to a standard size of 224×224, ensuring uniformity for further processing or analysis. Figure 2 displays example images from the original dataset alongside their aligned and tightly cropped ear images. Finally, we apply Contrast Limited Adaptive Histogram Equalization (CLAHE) to enhance the contrast of the images.

\begin{figure}[htb]
\centering
    \includegraphics[width=6cm]{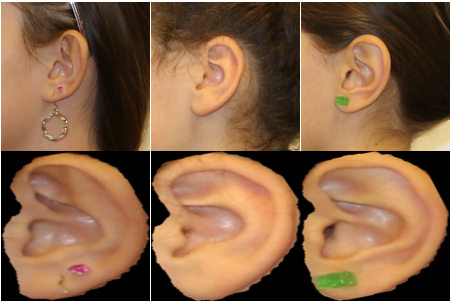}
    \caption{Example images from zoomed profile face and corresponding processed ear images}
    \label{logo}
\end{figure}

\subsection{Training and Evaluation Procedure}
Our ear recognition system designed for children's ear images using pre-trained VGG16 and MobileNet models for feature extraction. Initially, we load the VGG16 model without its top layer and freeze its weights to prevent retraining. We configure the data generators to rescale images to a [0, 1] range and compile the model using the RMSprop optimizer and categorical cross-entropy loss. In a parallel scenario, we employ the MobileNet model, excluding its classification layer while retaining its convolutional base, and add custom layers including a GlobalAveragePooling2D layer, a dense layer with ReLU activation, and dropout regularization to prevent overfitting. Both models are compiled with RMSprop, and categorical cross-entropy and include callbacks for early stopping, model checkpointing, and dynamic learning rate adjustment. Using around 80\% of child ear images for training, we preprocess the data and normalize pixel values. For testing, we preprocess the remaining dataset and load the pre-trained MobileNet and VGG16 models, extracting features from specific layers of each model. Feature vectors are refined using an ExtraTreesClassifier to identify the most informative features and reduce dimensionality, followed by concatenation into a unified feature vector for each image. Applying t-SNE for further dimensionality reduction, we calculate Euclidean distances between ear image pairs to evaluate the system's accuracy. Matching results are classified into True Accepts (TA), False Rejects (FR), False Accepts (FA), and True Rejects (TR), and we compute the True Acceptance Rate (TAR) and False Acceptance Rate (FAR) to assess system performance. This approach demonstrates a robust method for enhancing ear recognition accuracy in biometric applications.

\subsection{Results and Discussions}

\begin{figure}[htb]
\centering
    \includegraphics[width=6cm]{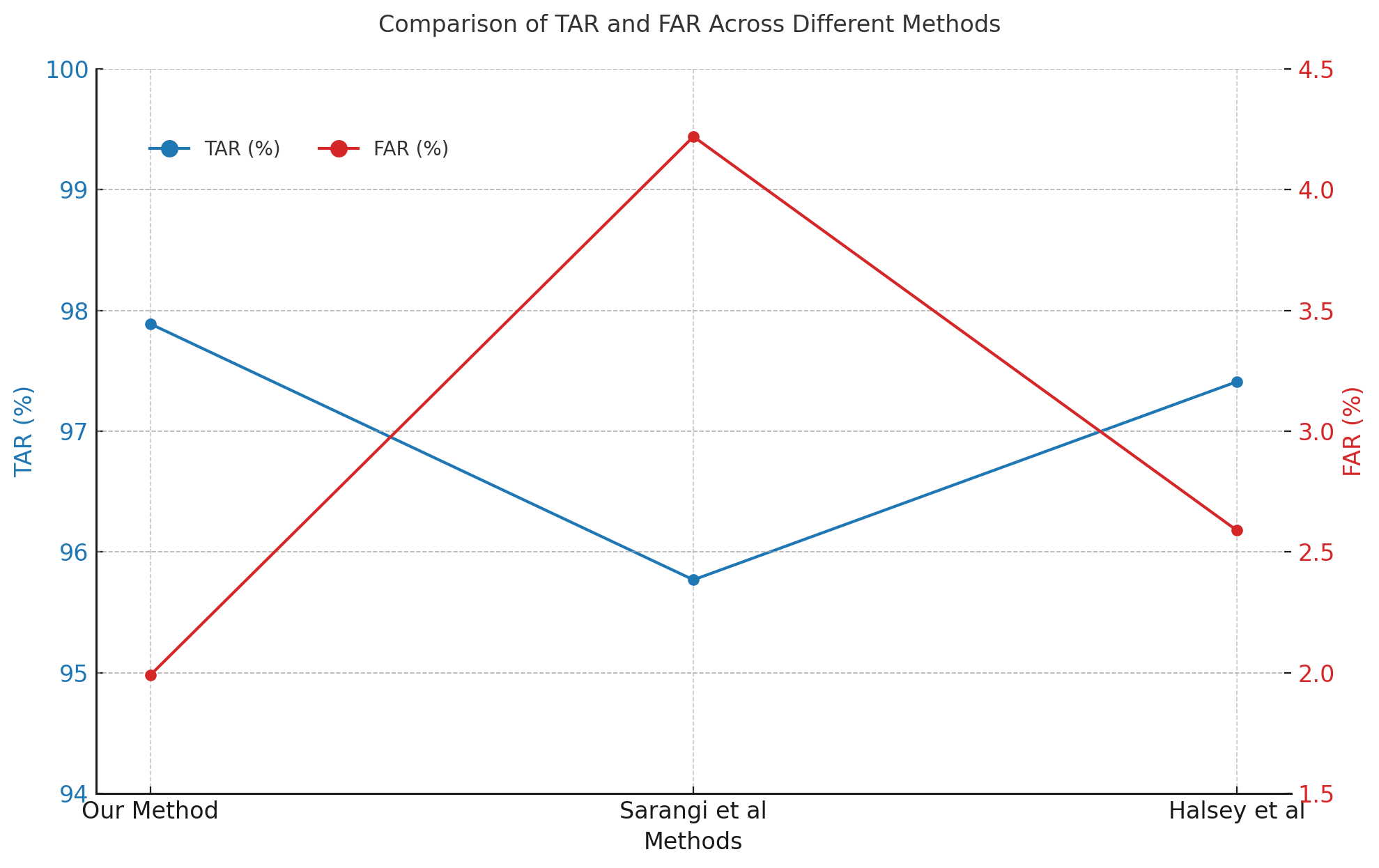}
    \caption{Performance Comparison of TAR and FAR Across Different Methods for IITD dataset}
    \label{logo}
\end{figure}

We evaluate the efficiency of our ensemble method using the IITD dataset to assess its performance. We train our model on 80\% of the IITD dataset and then calculate the True Acceptance Rate (TAR) and False Acceptance Rate (FAR). Our method achieves a TAR of 97.89\% while maintaining a FAR of 1.99\%. We have found that most authors report results using classification accuracy, while only a few provide recognition accuracy. Recognition involves comparing an image with a previously enrolled template. Conversely, classification categorizes individuals into predefined groups based on features, which can result in misclassifying an unenrolled person as an enrolled individual. Therefore, classification is not representative of real-life implementation, whereas recognition is. Figure 3 shows the comparison of our method with other methods that report recognition accuracy. The figure clearly demonstrates that our method outperforms the other methods Sarangi et al. \cite{AB44} and Hansley et al. \cite{AB45} in terms of recognition accuracy, achieving the highest TAR and the lowest FAR.

\begin{figure}[htb]
\centering
    \includegraphics[width=6cm]{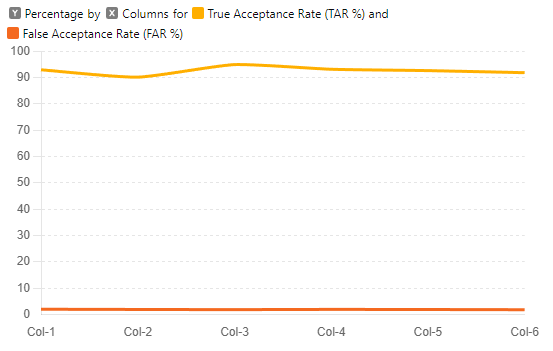}
    \caption{Performance evaluation of our ensemble method (combination of VGG16 and MobileNet) across six collections (Col-1 to Col-6) of our child dataset}
    \label{logo}
\end{figure}

For our child dataset, we train six distinct data sets independently using the ensemble approach with the VGG16 and MobileNet models and in this case, we are getting more than 90\%Tar for all collections keeping the FAR around 2\%. Figure 4 displays the outcome of our use of the ensemble model. During both training and testing, we observed that our Collection 2 data exhibited the lowest quality compared to other datasets. 

\begin{figure}[htb]
\centering
    \includegraphics[width=5cm]{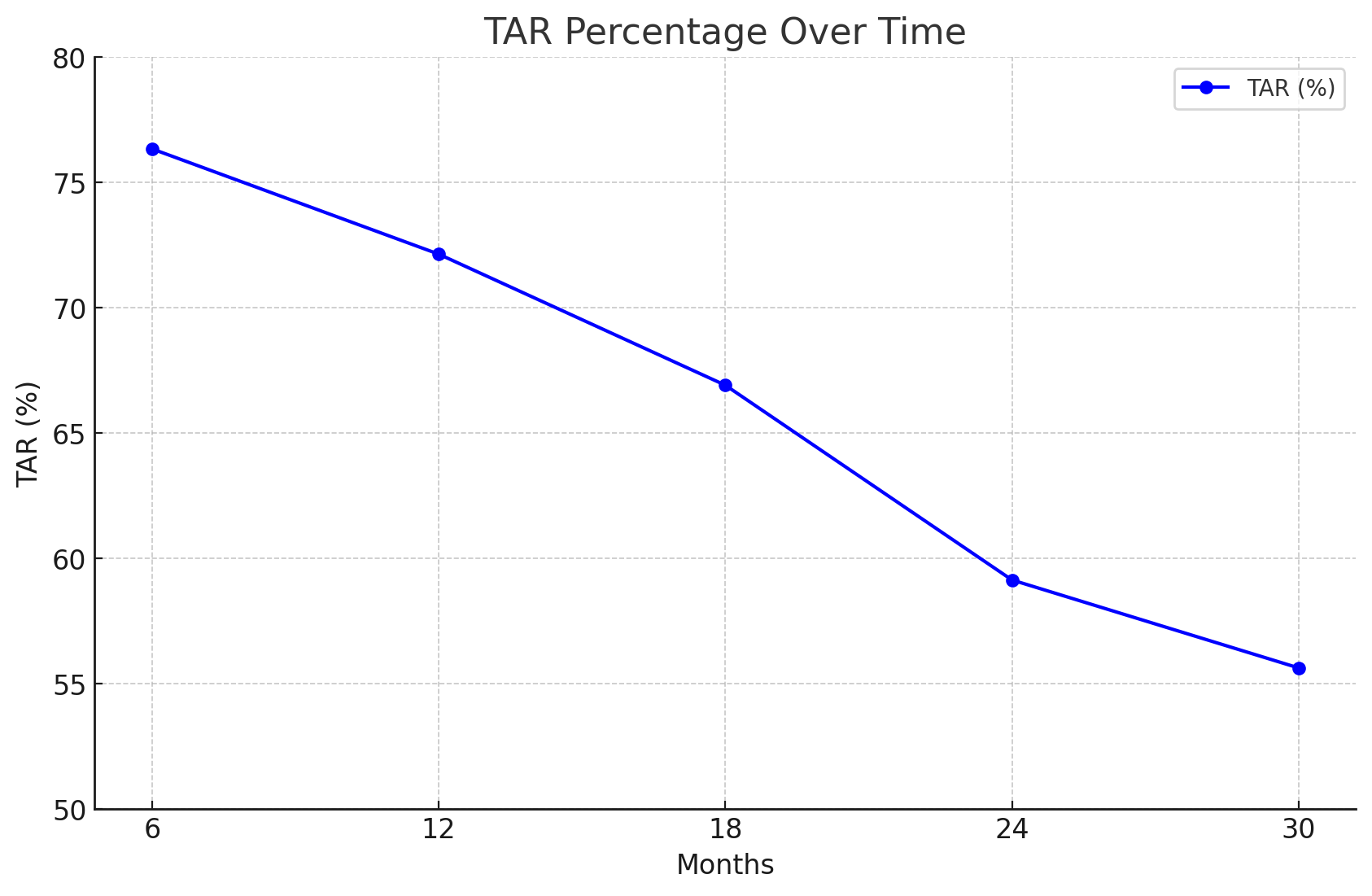}
    \caption{TAR\% at 2.0\% FAR for increasing time gaps between enrollment and verification}
    \label{logo}
\end{figure}

After applying our ensemble method independently to the six different collections, we then compared the accuracy across all the collections. We compared collection-1 against collections 2 through 6, with the results illustrated in Figure 5. The figure reveals that ear recognition for children does not remain effective over time. Although we achieve good accuracy within individual collections, the accuracy significantly drops when comparing one collection to another. While the algorithm performs exceptionally well on the IITD dataset, which consists of adult data, it fails to maintain performance over a 36-month period for children, with True Acceptance Rates (TAR) ranging from 55\% to 76\%. Previous research on child ear recognition has identified that the ear's structure undergoes rapid changes from 4 months to 8 years of age, leading to poor recognition results in our child dataset, as many subjects fall within this age range.

\begin{figure}[htb]
\centering
    \includegraphics[width=5cm]{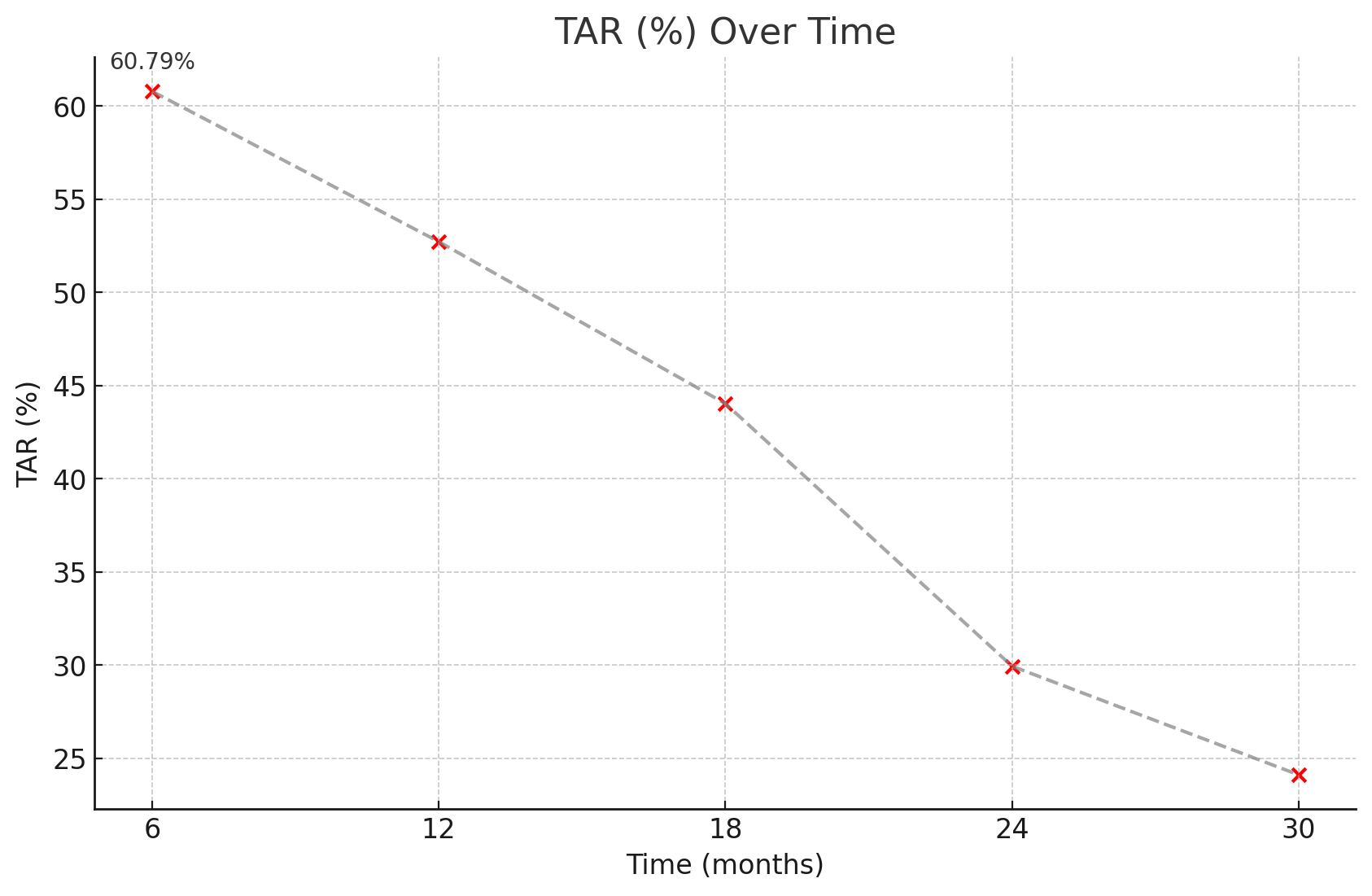}
    \caption{TAR\% at 2.0\% FAR for increasing time gaps between enrollment and verification (children age less than or equal to 8yeras)}
    \label{logo}
\end{figure}

\begin{figure}[htb]
\centering
    \includegraphics[width=5cm]{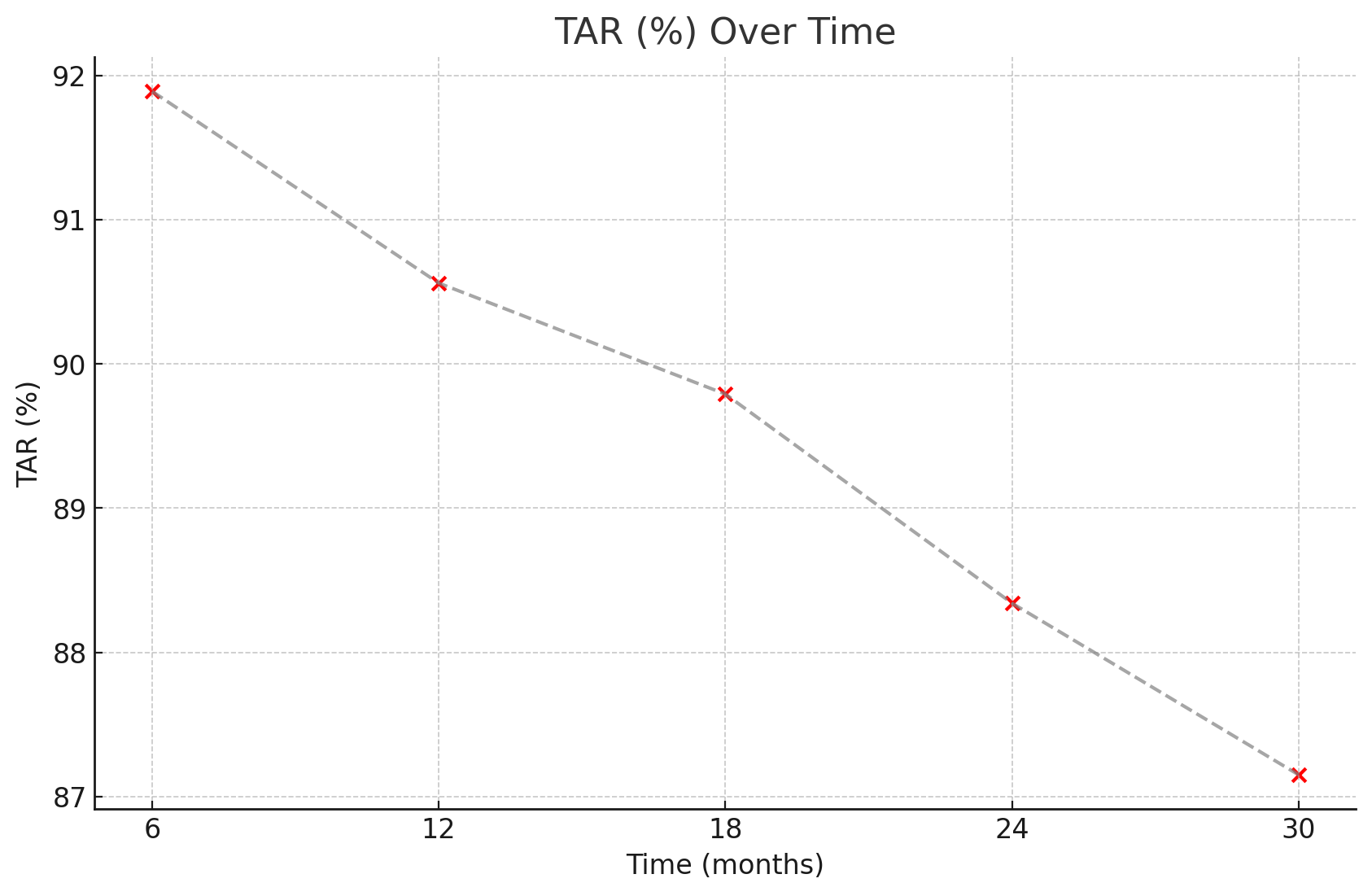}
    \caption{TAR\% at 2.0\% FAR for increasing time gaps between enrollment and verification (children age greater than  8yeras)}
    \label{logo}
\end{figure}

According to previous literature, the structure of a child's ear changes rapidly between 6 months and 8 years of age. Therefore, we separated the data into two groups: child ear data for those less than 8 years old and those greater than 8 years old. We then applied our method to calculate the accuracy. Our findings show that the accuracy is significantly lower for the dataset where test images are of children under 8 years old, supporting previous findings. Conversely, we achieve reasonable accuracy for the dataset where test images are of children above 8 years old. Figures 6 and 7 illustrate the TAR while maintaining the FAR around 2\%.

\section{Conclusion}
This study presents a comprehensive evaluation of ear recognition as a biometric modality with a particular focus on its application to children, addressing the unique challenges posed by the rapid structural changes in the ear during early childhood development. We introduced a longitudinal dataset collected from children aged 4 to 14 years over a period of 2.5 years and developed a deep learning based approach that combines VGG16 and MobileNet models to enhance recognition accuracy.

Our experiments revealed that while the ensemble method achieves a high True Acceptance Rate (TAR) exceeding 90\% within individual collections, its performance diminishes when comparing across different collections over time, reflecting the dynamic nature of ear growth in children. Specifically, the TAR for children (3years to 18years) ranged between 55\% to 76\% over a 30-month period where , compared to a TAR of 97.89\% achieved with an adult IID dataset. These findings underscore the challenges inherent in child ear recognition, which is compounded by significant changes in ear structure up to the age of eight.

In conclusion, while ear biometrics offers numerous advantages such as reduced invasiveness and enhanced privacy, its application in children remains challenging due to the rapid and significant developmental changes. Future work should focus on developing adaptive models that can account for and adjust to these changes over time, as well as exploring the integration of additional biometric modalities to improve robustness. This study lays the groundwork for further exploration and refinement of ear recognition systems, particularly for dynamic demographics such as children.

\bibliography{lniguide}

\end{document}